\documentclass[a4paper,UKenglish,nolinenumbers,cleveref,autoref,thm-restate]{socg-lipics-v2021}
\usepackage[T1]{fontenc}
\usepackage[utf8]{inputenc}

\usepackage{amsmath, mathtools, amssymb, amsfonts, amsthm}
\usepackage{graphicx}
\usepackage{tikz, tikz-cd}
\usetikzlibrary{matrix, automata, arrows, positioning, calc, shapes.geometric, arrows.meta}
\usepackage{pgfplots, pgfplotstable}
\usepackage{subcaption}
\usepackage{algorithm, algpseudocode}
\usepackage{booktabs, multicol}
\usepackage{microtype}
\usepackage{hyperref, url}
\usepackage{cleveref}
\usepackage{xcolor}
\usepackage{booktabs}
\usepackage{siunitx}
\usepackage{colortbl} % optional for shading
\usepackage[numbers,sort&compress]{natbib}
\usepackage{lineno}

\newcommand{\nedges}{\mathcal{N}_T}
\newcommand{\tim}{\mathsf{\tau}}

\title{Filtration-Based Representation Learning for Temporal Graphs}
\author{ Samrik Chowdhury}{Indian Institute of Science Education and
  Research Tirupati, India.}{}{}{}{} % Leave empty for double-blind
\author{Siddharth Pritam}{Chennai Mathematical Institute, India.}{spritam@cmi.ac.in}{}{}
\author{Rohit Roy}{Chennai Mathematical Institute, India.}{}{}{}
\author{Madhav Cherupilil Sajeev}{Institut Polytechnique de Paris, France.}{}{}{}{}

\authorrunning{Samrik Chowdhury et al.}     % Leave empty in anonymous mode
\Copyright{Samrik Chowdhury et al.}         % Leave empty in anonymous submissions
 %{Replace with valid ACM 2012 CCS classification}
\begin{document}
\maketitle
\nolinenumbers

% Paper content begins here

\begin{abstract}
%Temporal graphs effectively model dynamical systems via timestamped interactions. While many methods study temporal evolution by comparing graph snapshots, quantifying similarity between temporal graphs as a whole remains an open problem. 
In this work, we introduce a filtration on temporal graphs based on
$\delta$-temporal motifs (recurrent subgraphs), yielding a multi-scale
representation of temporal structure. %Filtrations, central to topological data analysis through persistent homology, provide a general framework for multi-scale analysis. Recent advances in graph learning show that filtrations extend naturally beyond Betti numbers, supporting multi-scale feature extraction and enabling structural graph kernels—such as the Weisfeiler--Lehman kernel—to be enriched through filtered representations.
Our temporal filtration allows tools developed for
filtered static graphs, including persistent homology and recent graph
filtration kernels, to be applied directly to temporal graph analysis.
We demonstrate the effectiveness of this approach on temporal graph
classification tasks.

\end{abstract}

% keywords can be removed

\section{Introduction}
A nested sequence of spaces
\((X_1 \subseteq X_2 \subseteq \cdots \subseteq X_n)\) is referred to as a
\textit{filtration} and remains a simple yet highly effective idea for
studying an object or space across multiple scales. It captures how features evolve from
local to global structure. Persistent homology (PH) is a central example:
by tracking homological features through a filtration, it yields a richer
and more nuanced understanding of a space's topology. Recently, this idea
of strengthening a discriminating invariant via filtration has been
extended to graph learning through \emph{graph filtration
kernels}~\cite{Schulz_2022}. Schulz et al.~\cite{Schulz_2022} show that
filtered graph kernels are strictly more expressive than static graph
kernels, precisely because filtrations provide a unified yet fine-grained,
and orderly representation of the underlying structure. This added
expressiveness is especially relevant given that determining whether two
graphs are isomorphic remains computationally
challenging~\cite{babai2016gi}.

Given the significance and availability of general tools such as PH and
filtered graph kernels—which rely on filtered representations—it is
advantageous to analyze an object through a suitable filtration. In this
article, we extend this perspective to the study of \emph{temporal
graphs}~\cite{Holme2012}, where edges carry timestamps representing time-specific
interactions. We propose a method that captures temporal connectivity
without aggregating data into discrete snapshots, enabling each temporal
graph to be treated as a single coherent entity. This provides a
filtration-based alternative to snapshot-based
methods~\cite{Araujo2014, Dunlavy2011, tinarrage2023TDANetVis,
myers2023temporal, shamsi2024graphpulse}, which often obscure the finer
temporal structure.

\subparagraph{Our Approach:} We use $\delta$-temporal motifs, introduced in~\cite{paranjape2017motifs}—small
connected temporal subgraphs whose edges occur within a $\delta$-time
window (see Section~\ref{sec:prelem} for the precise definition)—to
define local distances between nodes. These distances induce a filtration that captures both the frequency and timing of
interactions, formalized as the \emph{average filtration} in
Section~\ref{sec:temp_filtration}. The use of \(\delta\)-temporal motifs is motivated by~\cite{paranjape2017motifs},
which shows experimentally that the distribution of various
\(\delta\)-motifs differs significantly across classes of temporal graphs.
More generally, local
neighborhood structures are central to studying isomorphism properties
of static graphs, as exemplified by the Weisfeiler--Lehman
framework~\cite{shervashidze2011,weisfeiler1968}.

We then use the resulting average filtration in two ways for temporal graph classification: first, by applying \textit{persistent homology} to extract topological features such as connectivity and cycles, yielding a
general-purpose descriptor (a \emph{persistence diagram}, see Section~\ref{sec:prelem}); and second, by employing the recently introduced graph filtration kernels~\cite{Schulz_2022}. Both approaches produce kernel matrices that are subsequently used to train an SVM classifier.

\subparagraph{Our Contribution: }
Our main contributions are as follows:
\begin{enumerate}
    \item We introduce a filtration-based     framework for representing and analyzing
     temporal graphs without relying on snapshot aggregation. Our method
     naturally handles temporal graphs of varying sizes and offers broader
     applicability than existing approaches that depend on node-level labels.

    \item On the theoretical side, we establish the stability of our filtration
    under standard temporal graph reference models (see Section~\ref{sec:Stability}). We also provide an efficient algorithm for
    computing the average filtration with time complexity
    \(\mathcal{O}(|E| \times d_{\max})\) and memory usage comparable to that
    of an aggregated graph, offering substantial space savings relative to
    storing full temporal data. Here, \(|E|\) is the number of temporal edges
    and \(d_{\max}\) is the maximum temporal degree.
    
    \item We extensively benchmark our framework for temporal graph classification, achieving over \(92\%\) accuracy in all settings and \(100\%\) in several cases (see Section~\ref{sec:experiments}), all without requiring node-level labels.

\end{enumerate}

\subparagraph{\textbf{Related work:}}
Temporal graphs have been widely used in applications ranging from
ecological systems to human close-range interactions, collaboration
networks, and biological signaling
networks~\cite{Holme2012, eagle2009inferring, cattuto2010dynamics,
barrat2013empirical, stopczynski2014measuring, BravoHermsdorff2019}.
Characterizing and classifying temporal graphs is an important and active research
area, with approaches spanning kernel methods~\cite{Oettershagen2020}, embedding distances~\cite{DallAmico2024},
temporal motifs~\cite{paranjape2017motifs, Kovanen2011}, and neural networks~\cite{Sato2019DyANE, Rossi2020TGN}.

Numerous metrics and tools exist for assessing
structural properties of static graphs, and while some—such as path
length, centrality, and betweenness—have been extended to temporal
graphs~\cite{Holme2012}, comparatively few methods analyze temporal
graphs as holistic objects. Dall’Amico et al.~\cite{DallAmico2024} address this gap by defining a
metric on the space of temporal graphs, allowing each temporal graph to
be treated as a single entity. Our work extends this perspective by
introducing a more general, filtration-based representation of temporal
graphs, providing a more fine-grained framework for their comparison and
analysis.

Ye et al.~\cite{stabledistance} also define filtrations for dynamic
graphs and use the resulting persistence diagrams for classification.
However, their model relies on time-varying edge weights to determine
filtration values, yet—somewhat surprisingly—most of their experiments
use unweighted graphs. This makes their approach appear less directly aligned with standard temporal networks, which are typically modeled as contact sequences. In the context of applying persistent homology to temporal data, Tinarrage et
al.~\cite{tinarrage2023TDANetVis} and Myers et~\cite{myers2023temporal}
employ zigzag PH to analyze temporal networks, with the former using
discrete time-window resolutions for visualization. While effective,
zigzag persistence is computationally expensive and highlights the
difficulty of capturing temporal structure without aggregation, as well
as the challenge of selecting meaningful temporal
resolutions~\cite{tinarrage2023TDANetVis}.

The work by Oettershagen et al.~\cite{Oettershagen2020} introduces three distinct techniques for mapping temporal graphs to static graphs, thereby enabling the application of conventional static graph kernels. Wang ~\cite{wang2018time} explore the classification of temporal graphs where both vertex and edge sets evolve over time.
Tu et al.~\cite{Tu2018NetworkCI} leverage temporal motifs, while Wang et al.~\cite{Wang2020TimeVariant} propose time-variant graph classification via \emph{graph-shapelet patterns}
extracted from sequences of snapshot-based graph statistics. %, a method that is somewhat computationally heavy and tailored to specific tasks.

\section{Preliminaries}\label{sec:prelem}
In this section, we briefly recall basic notions related to temporal networks, persistent homology and kernel methods. Readers may refer to~\cite{Edelsbrunner2010, Hatcher2002,Holme2012, Scholkopf2002} for more details.

\subsection{Temporal Network}
%A temporal network is a dynamic variant of the static networks in which the edges (interactions or connections) are associated with time stamps (labels). These dynamic edges can change over time, which is essential for modeling systems where the timing of interactions is crucial.
%Below, we recall useful definitions, for more details readers can refer to~\cite{Holme2012}.\\

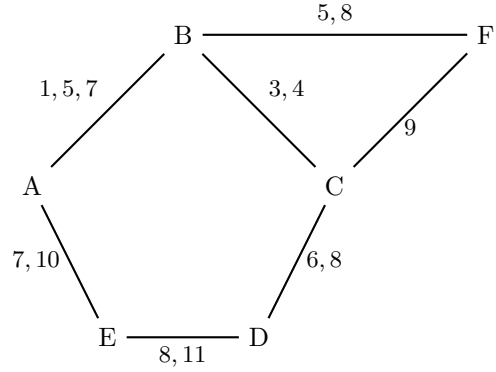
\begin{figure}[h!]
\centering

\begin{tikzpicture}[-, >=latex, node distance=2cm, thick, scale=1]
        % Nodes
        \node (A) at (0, 2) {A};
        \node (B) at (2, 4) {B};
        \node (C) at (4, 2) {C};
        \node (D) at (3, 0) {D};
        \node (E) at (1, 0) {E};
        \node (F) at (6, 4) {F};

        % Temporal edges with timestamps
        \path[every node/.style={font=\sffamily\small}]
            (A) edge node[above left] {$1,5,7$} (B)
            (B) edge node[above right] {$3,4$} (C)
            (C) edge node[right] {$6,8$} (D)
            (D) edge node[below ] {$8,11$} (E)
            (E) edge node[left] {$7,10$} (A)
            (B) edge node[above] {$5,8$} (F)
            (F) edge node[below] {$9$} (C);

        % Title
        %\node[below of=F, node distance=2cm] {Temporal Graph};
    \end{tikzpicture}
\caption{Multiple time labels are separated by commas.}
\label{fig:mulit_temporal_graph}
\end{figure}

\subparagraph{\bf{Temporal Graphs:}} A \textbf{temporal graph} is defined as a tuple \( T = (V, E) \), where \( V \) represents a set of vertices (nodes), and \( E \) is a set of directed or undirected \textit{temporal edges}. Each temporal edge \( e := (u, v, t) \) connects vertices \( u \) and \( v \) and is active only at a specific time \( t \). Alternatively, a temporal graph is as a sequence of contacts (interactions), where each temporal edge is represented as a contact \( (u, v, t) \). If there is only a single temporal edge between any two nodes, the graph is referred to as a \textbf{single-labeled temporal graph}. When multiple temporal edges exist between two nodes, such as \( (u, v, t_1) \) and \( (u, v, t_2) \) etc., the graph is called a \textbf{multi-labeled temporal graph}. Collectively, these temporal edges are referred to as the edges between \( u \) and \( v \). Furthermore, if all the edges (interactions) have a time duration \([t_1, t_2]\), the graph is known as an \textbf{interval temporal graph} and the temporal edge is denoted as  \( e = (u, v, [t_1, t_2]) \). The \textbf{temporal degree} of a vertex $u$ is the number of temporal edges connected to it, denoted by $td_u$. %In an undirected temporal graph, $td_u = |(v,v',t)|$ such that $v = u$ or $v' = u$. 
See Figure~\ref{fig:mulit_temporal_graph} for an example of multi-labeled temporal graph.

% Temporal Graph Drawing (optional)

%\begin{figure}[h!]
%\centering
%\begin{tikzpicture}[-, >=latex, node distance=2cm, thick]
%        % Nodes
%        %\node (A) at (0, 2) {A};
%        \node (B) at (2, 4) {B};
%        \node (C) at (4, 2) {C};
%        %\node (D) at (3, 0) {D};
%        %\node (E) at (1, 0) {E};
%        \node (F) at (6, 3) {F};
%
%        % Temporal edges with timestamps
%        \path[every node/.style={font=\sffamily\small}]
%            %(A) edge node[above left] {1/5/7} (B)
%            (B) edge node[above right] {4} (C)
%            %(C) edge node[right] {6/8} (D)
%            %(D) edge node[below ] {8/11} (E)
%            %(E) edge node[left] {7/10} (A)
%            (B) edge node[above] {5/8} (F)
%            (F) edge node[below] {9} (C);
%
%    \end{tikzpicture}
%\caption{Example of a multi-labeled temporal graph.}
%\label{fig:temporal_graph}
%\end{figure}

For simplicity, this article focuses on undirected temporal graphs; however, all concepts and methods can naturally extend to directed temporal graphs. When the context is clear, a temporal edge is referred to simply as an \textit{edge}. By ignoring timestamps and duplicate edges, a temporal graph induces a simple static graph, commonly referred to as the \textbf{aggregate graph} \( G \) of \( T \). In \( G \), a pair \( (u, v) \) is an edge if and only if there exists a temporal edge \( (u, v, t) \) in \( T \).\\

%\subparagraph{Temporal Reachability:} One of the fundamental concepts in temporal networks is \textit{temporal reachability}. A vertex \( v \) is said to be \textbf{reachable} from another vertex \( u \) if there exists a temporal path \(\{ (u, v_1, t_1), (v_1, v_2, t_2), \ldots, (v_n, v, t_n) \} \) such that \( t_i \leq t_{i+1} \) for all \( i \in [1, n-1] \). Unlike static networks, where paths are defined by the presence of edges regardless of time, temporal reachability depends on the timing of interactions. This distinction is crucial for modeling processes that rely on the timing and sequence of interactions, such as the spread of information or diseases.

%\begin{definition} \label{def:dtg}
%A directed temporal graph $G$ consists of a finite set $V$ of nodes and a finite set $E \subseteq V \times V \times \mathbb{N} $ 
%%of ordered pairs of nodes 
%of interactions. An interaction $e \in E$
%is represented by a three-tuple $e = (u, v, t)$, in which $u$ is the source node, $v$
%is the target node and $t$ is the initiation time of the interaction.
%\end{definition}

%\begin{definition}\label{def:timerespectinginter}
%Let $e_i = (u_i, v_i, t_i)$ and $e_j = (u_j, v_j , t_j)$ be two interactions in a directed temporal graph $G$. Given some threshold $d$, the interactions are \textit{timerespecting} if they are adjacent edges and:
%
%\begin{enumerate}
%    \item $0 \leq t_j - t_i \leq d; \text{ if } v_i = u_j,$
%    \item $0 \leq t_i - t_j \leq d; \text{ if } u_i = v_j,$
%    \item $0 \leq |t_j - t_i| \leq d; \text{ otherwise}$.
%\end{enumerate} 
%\end{definition}

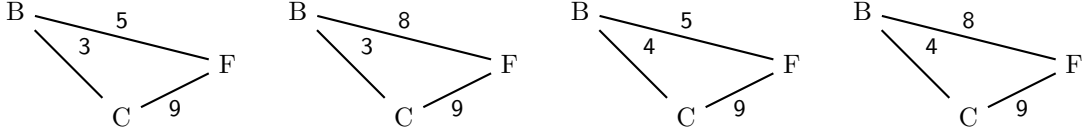
\begin{figure}[h!]
%\centering
\begin{minipage}{0.2\textwidth}
\begin{tikzpicture}[scale=0.7, -, >=latex, node distance=2cm, thick]
        % Nodes
        %\node (A) at (0, 2) {A};
        \node (B) at (2, 4) {B};
        \node (C) at (4, 2) {C};
        %\node (D) at (3, 0) {D};
        %\node (E) at (1, 0) {E};
        \node (F) at (6, 3) {F};

        % Temporal edges with timestamps
        \path[every node/.style={font=\sffamily\small}]
            %(A) edge node[above left] {1/5/7} (B)
            (B) edge node[above right] {3} (C)
            %(C) edge node[right] {6/8} (D)
            %(D) edge node[below ] {8/11} (E)
            %(E) edge node[left] {7/10} (A)
            (B) edge node[above] {5} (F)
            (F) edge node[below] {9} (C);

    \end{tikzpicture}
\end{minipage}
\hfill
\begin{minipage}{0.2\textwidth}
\begin{tikzpicture}[scale=0.7, -, >=latex, node distance=2cm, thick]
        % Nodes
        %\node (A) at (0, 2) {A};
        \node (B) at (2, 4) {B};
        \node (C) at (4, 2) {C};
        %\node (D) at (3, 0) {D};
        %\node (E) at (1, 0) {E};
        \node (F) at (6, 3) {F};

        % Temporal edges with timestamps
        \path[every node/.style={font=\sffamily\small}]
            %(A) edge node[above left] {1/5/7} (B)
            (B) edge node[above right] {3} (C)
            %(C) edge node[right] {6/8} (D)
            %(D) edge node[below ] {8/11} (E)
            %(E) edge node[left] {7/10} (A)
            (B) edge node[above] {8} (F)
            (F) edge node[below] {9} (C);

    \end{tikzpicture}
\end{minipage}
\hfill
\begin{minipage}{0.2\textwidth}
\begin{tikzpicture}[scale=0.7, -, >=latex, node distance=2cm, thick]
        % Nodes
        %\node (A) at (0, 2) {A};
        \node (B) at (2, 4) {B};
        \node (C) at (4, 2) {C};
        %\node (D) at (3, 0) {D};
        %\node (E) at (1, 0) {E};
        \node (F) at (6, 3) {F};

        % Temporal edges with timestamps
        \path[every node/.style={font=\sffamily\small}]
            %(A) edge node[above left] {1/5/7} (B)
            (B) edge node[above right] {4} (C)
            %(C) edge node[right] {6/8} (D)
            %(D) edge node[below ] {8/11} (E)
            %(E) edge node[left] {7/10} (A)
            (B) edge node[above] {5} (F)
            (F) edge node[below] {9} (C);

    \end{tikzpicture}
\end{minipage}
\hfill
\begin{minipage}{0.2\textwidth}
\begin{tikzpicture}[scale=0.7, -, >=latex, node distance=2cm, thick]
        % Nodes
        %\node (A) at (0, 2) {A};
        \node (B) at (2, 4) {B};
        \node (C) at (4, 2) {C};
        %\node (D) at (3, 0) {D};
        %\node (E) at (1, 0) {E};
        \node (F) at (6, 3) {F};

        % Temporal edges with timestamps
        \path[every node/.style={font=\sffamily\small}]
            %(A) edge node[above left] {1/5/7} (B)
            (B) edge node[above right] {4} (C)
            %(C) edge node[right] {6/8} (D)
            %(D) edge node[below ] {8/11} (E)
            %(E) edge node[left] {7/10} (A)
            (B) edge node[above] {8} (F)
            (F) edge node[below] {9} (C);

    \end{tikzpicture}
\end{minipage}
\caption{All four possible 3-node 3-edge motifs of Figure~\ref{fig:mulit_temporal_graph} are depicted here. Note that the temporal edges between $B$ and $C$ and $B$ and $F$ differ in their respective timestamps. The two left motifs satisfy $\delta = 6$, while the two right motifs satisfy $\delta = 5$.
}
\label{fig:four_motifs}
\end{figure}

\subparagraph{\bf{$\delta$-Temporal Motifs:}}\label{def:temporal_motif} % $\delta$-Temporal motifs, introduced in~\cite{paranjape2017motifs}, extend the concept of motifs (recurring subgraphs) from static to temporal graphs. 
We recall the definition of $\delta$-temporal motifs from~\cite{paranjape2017motifs} in the context of undirected temporal graphs. A $k$-node, $l$-edge, \textbf{$\delta$-temporal motif} is a sequence of $l$ edges, $M = \{(u_1, v_1, t_1), (u_2, v_2, t_2), \dots, (u_l, v_l, t_l)\}$ such that the edges are time-ordered within a $\delta$ duration, i.e., 
\(
t_1 < t_2 < \dots < t_l \quad \text{and} \quad t_l - t_1 \leq \delta,
\)
and the induced aggregate graph from the edges is connected and has $k$ nodes.
Note that with this general definition, multiple edges between the same pair of nodes may appear in the motif \( M \). However, we restrict our attention to the case where for any pair of temporal edges \( (u_i, v_i, t_i), (u_j, v_j, t_j) \) in \( M \), it is not true that \( u_i = u_j \) and \( v_i = v_j \)\footnote{This restriction is necessary to define a filtration of simplicial complexes.}. See Figure~\ref{fig:four_motifs} for examples.

\subsection{Topological Data Analysis}

\subparagraph{\bf{Simplicial Complex:}}
A \textbf{$k$-simplex} on a vertex set $V$ is any subset
$\sigma \subseteq V$ with $|\sigma| = k+1$, and every nonempty
$\tau \subseteq \sigma$ is called a \textbf{face} of $\sigma$. An
\textbf{abstract simplicial complex} $K$ is a collection of finite
subsets of a vertex set such that
\(
\sigma \in K \ \text{and}\ \tau \subseteq \sigma \;\Rightarrow\; \tau \in K.
\)
Thus $K$ is closed under taking faces. A \textbf{subcomplex} is any
subcollection $L \subseteq K$ that satisfies the same closure property. We will refer to an \textit{abstract simplicial complex} as simply a \textit{simplicial complex} or just a \textit{complex}.
A complex $K$ is a \textbf{flag} or \textbf{clique} complex if, whenever a subset of its vertices forms a clique (i.e., any pair of vertices is connected by an edge), they span a simplex. It follows that the full structure of $K$ is determined by its 1-skeleton (or graph), which we denote by $G$.

\iffalse
\subparagraph{\bf{Homology:}} \textit{Homology} is a tool from algebraic topology to quantify the number of $k$-dimensional topological features (or holes) in a topological space, such as a simplicial complex. For instance, $H_0$ the zero degree homology class describes the number of connected components, $H_1$ the one degree homology class describes the number of loops, and $H_2$, the two dimensional homology class quantifies voids or cavities. The ranks of these homology groups are referred to as \textbf{Betti numbers}. In particular, the Betti number $\beta_ k$ corresponds to the number of $k$-dimensional holes. In Figure~\ref{fig:simpcompexp} the Betti numbers are as follows, $\beta_0 =1, \beta_1 = 1, \text{ and } \forall k \geq 2, \beta_k = 0$.
\fi
\subparagraph{\bf{Filtration:}}
A \textbf{sequence} of simplicial complexes $\mathcal{F}$: $\{K_1 \hookrightarrow K_2 \hookrightarrow \cdots \hookrightarrow K_m \}$ connected through inclusion maps is called a \textbf{filtration}. A filtration is called a \textbf{flag filtration} if all the simplicial complexes $K_i$ are flag complexes. Given a weighted graph $G = (V, E, w : E \rightarrow \mathbb{R})$, a flag filtration $F_G$ can be defined by assigning the maximum weight of edges in a simplex (clique) as its filtration value. Flag filtrations are among the most common types of filtrations used in TDA applications. %The concept of filtration is used to analyze data across multiple scales. 
As the filtration \textit{parameter} increases, new simplices are added, allowing us to track topological features, such as Betti numbers, emerge and disappear across different scales. The next paragraph formalizes this intuition.

%In a standard pipeline of \textit{persistent homology} computation, a dataset, typically represented as a point cloud in a metric space or a filtered graph, is used to first build a filtered simplicial complex which is used to construct a \textit{boundary matrix} which is then finally reduced in a special form to read off the persistent homology. 

% A common method for constructing such filtered simplicial complex is the \textit{Vietoris-Rips} complex. This complex is formed by connecting data points that lie within a certain distance from each other, progressively increasing the complexity of the structure as the distance threshold grows. 

\subparagraph{\bf{Persistent Homology:}} If we compute the homology groups of all the $K_i$, we obtain the sequence $\mathcal{P}(\mathcal{F})$: $\{H_p(K_1) \xrightarrow{*} H_p(K_2) \xrightarrow{*} \cdots \xrightarrow{*} H_p(K_m)\}$. Here $H_p()$ denotes the homology group of dimension $p$ with coefficients from a field $\mathbb{F}$, and $\xrightarrow{*}$ is the homomorphism induced by the inclusion map. $\mathcal{P}(\mathcal{F})$ forms a sequence of vector spaces connected through the homomorphisms, called a \textbf{persistence module}. %More formally, a \textit{persistence module} $\mathbb{V}$ is a sequence of vector spaces $\{V_1 \rightarrow V_2 \rightarrow V_3 \rightarrow \cdots \rightarrow V_m\}$ connected by homomorphisms. A persistence module arising from a sequence of simplicial complexes captures the evolution of the topology across the sequence.

Any persistence module can be \textit{decomposed} into a collection of simpler interval modules of the form $[i, j)$~\cite{zomorodian2005computing}. The multiset of all intervals $[i, j)$ in this decomposition is called the \textbf{persistence diagram} of the persistence module. An interval of the form $[i, j)$ in the persistence diagram of $\mathcal{P}(\mathcal{F})$ corresponds to a homological feature (a ‘cycle’) that appears at $i$ and disappears at $j$. The persistence diagram (PD) completely characterizes the persistence module, providing a bijective correspondence between the PD and the equivalence class (isomorphic) of the persistence module~\cite{Edelsbrunner2010, zomorodian2005computing}. 

%The \textit{bottleneck distance} is a metric to compare two persistence diagrams \( D_1 \) and \( D_2 \), which are multisets of points \((b, d)\) representing topological features. It measures the minimum cost to match points in \( D_1 \) and \( D_2 \), allowing matches with the diagonal \( \Delta \) for unmatched points. Formally, it is defined as:

%\[
%W_{\infty}(D_1, D_2) = \inf_{\gamma} \sup_{x \in D_1} \| x - \gamma(x) \|_{\infty},
%\]

%where \( \gamma \) is a bijection between \( D_1 \cup \Delta \) and \( D_2 \cup \Delta \), and \( \| x - \gamma(x) \|_{\infty} = \max(|b_1 - b_2|, |d_1 - d_2|) \). This distance captures the maximum discrepancy between matched points and is stable under small perturbations.

\subsection{PSS Kernel and Graph Filtration Kernel}

%\subparagraph{Kernel:}
%A \textbf{kernel} in machine learning and data analysis is a function that computes a similarity (inner product) between two data points in a potentially high-dimensional space without explicitly mapping the data to that space. % Kernels are foundational in algorithms like Support Vector Machines (SVMs), kernel PCA, and Gaussian Processes. They enable powerful transformations of data, making complex patterns more accessible for linear analysis.
%Given a set of points \( x \) and \( y \) in an input space \( X \), a \textbf{kernel} function \( \mathit{k}: X \times X \rightarrow \mathbb{R} \) is defined as: $\mathit{k}(x, y) = \langle \phi(x), \phi(y) \rangle,$ where \( \phi \) is a mapping from the input space \( X \) to a higher-dimensional feature space \( \mathcal{F} \). The function $\mathit{k}$ is \emph{symmetric} and \emph{positive semi-definite} (the corresponding Gram matrix). The kernel \( \mathit{k}(x, y) \) computes the inner product of \( \phi(x) \) and \( \phi(y) \) in \( \mathcal{F} \) without needing to explicitly compute \( \phi \), a property known as the \textit{kernel trick}. A very common kernel is the \textit{Gaussian Kernel}: \( \mathit{k}(x, y) = \exp\left(-\frac{\|x - y\|^2}{2\sigma^2}\right) \), which maps to an infinite-dimensional feature space.

\subparagraph{\bf{Kernel for Persistence Diagrams:}} A kernel for persistence diagrams quantifies the similarity between diagrams by computing a weighted sum of inner products of feature points. %\footnote{A direct measure of distance between persistence diagrams is the \textit{bottleneck distance}~\cite{cohen2007stability}. However, since the space of persistence diagrams is non-linear, kernel-based distances are more suitable for machine learning applications.}. 
A commonly used kernel for PDs is the Persistence Scale Space (PSS) Kernel.  %The PSS Kernel accomplishes this by mapping persistence points to a new space where their distances are well-defined and meaningful~\cite{kusano2016persistence}. 
The \textbf{Persistence Scale Space (PSS) Kernel}~\cite{reininghaus2015stable} is defined for two persistence diagrams \( D \) and \( D' \) as follows:
\[
K_{PSS}(D, D') = \frac{1}{8\pi \sigma^2}\sum_{(p \in D)} \sum_{(q \in D')} e^{\left(-\frac{1}{8\sigma} \left( \| p - q \|^2 \right) \right)} - e^{\left(-\frac{1}{8\sigma} \left( \| p - \bar{q} \|^2 \right) \right)},
\]
where \(\sigma\) is a bandwidth parameter and $\bar{q} = (b, a)$ is $q = (a, b)$ mirrored at the diagonal. The PSS Kernel is stable under small perturbations of the input, ensuring that small changes in the diagram do not result in drastic kernel value changes, which is essential for robust applications in noisy data. %The PSS Kernel enables the analysis and comparison of persistence diagrams using a continuous and differentiable kernel function, making it suitable for integration with machine learning tasks, especially in non-Euclidean data settings~\cite{reininghaus2015stable}.

\subparagraph{Graph Filtration Kernel: } A \textbf{graph filtration} is a nested sequence of graphs
\(G_1 \subseteq G_2 \subseteq \cdots \subseteq G_k\). For an
edge-weighted graph \(G = (V, E, w)\) with \(w : E \to \mathbb{R}_{>0}\),
a standard filtration is obtained by thresholding the edge weights:
\(
A(G):\quad G_1 \subseteq G_2 \subseteq \cdots \subseteq G_k = G,
\)
where each \(G_i = (V, E_i)\) is defined by
\(E_i = \{\, e \in E : w(e) \ge \alpha_i \,\}\) for thresholds
\(\alpha_1 \ge \alpha_2 \ge \cdots \ge \alpha_k = 0\)
(cf.~\cite{Schulz_2022}).Let $F$ be a finite set of graph features (e.g., WL labels), and fix
$f \in F$. The \emph{filtration histogram}
$\phi_A^f(G)\in\mathbb{R}^k$ records the number of occurrences of $f$ in
each filtration level $G_i$:
\[
\phi_A^f(G)_i=\#\{f \text{ in } G_i\},\qquad i=1,\dots,k.
\]
Let $A_\alpha=\{\alpha_1,\dots,\alpha_k\}$ with ground distance
$d_1(\alpha_i,\alpha_j)=|\alpha_i-\alpha_j|$, and let $W_{d_1}$ denote
the corresponding 1-Wasserstein distance. Using the $\ell_1$-normalized
histogram $\hat{\phi}_A^f(G)=\phi_A^f(G)/\|\phi_A^f(G)\|_1$, the
\emph{base filtration kernel} for feature $f$ and $\gamma>0$ is
\begin{equation}
k_f^A(G,G')
=\exp\!\Bigl(
-\gamma\, W_{d_1}\!\left(
\hat{\phi}_A^f(G),\,
\hat{\phi}_A^f(G')
\right)
\Bigr).
\label{eq:base-filtration-kernel}
\end{equation}
The \emph{graph filtration kernel} aggregates these base kernels using
the original histogram masses:
\begin{equation}
K^{\mathrm{Filt}}_{F,A}(G,G')
=\sum_{f\in F}
k_f^A(G,G')\,
\|\phi_A^f(G)\|_1\,
\|\phi_A^f(G')\|_1.
\label{eq:graph-filtration-kernel}
\end{equation}
Intuitively, this kernel measures similarity by comparing how features
appear and evolve across the filtration.%Following \cite{Schulz_2022}, the use of the Wasserstein distance with ground metric $d_1$ ensures that $k^A_f$ is positive semidefinite, and hence $K^{\mathrm{Filt}}_{F,A}$ defines a valid graph kernel. 
The filtration kernel is strictly more expressive than its base
feature map (e.g., WL subtree features) because it captures how features
evolve along the filtration. As shown in~\cite{Schulz_2022}, graphs that
standard histogram-based kernels cannot distinguish become separable when
their filtration histograms are compared via the Wasserstein distance.

\section{Temporal Filtrations}\label{sec:temp_filtration}
In this section, we introduce a simple method and an algorithm for constructing filtered (weighted) graphs from single-labeled, and multi-labeled temporal graphs. The filtered graph captures the evolution of fixed-size (\(3\)-node, \(2\)-edge) \( \delta \)-temporal motifs. %for increasing values of \( \delta > 0 \) in the given temporal graph. %The corresponding persistence diagram, computed by treating the filtered graph as the \(1\)-skeleton of a flag filtration, effectively captures the global `temporal topology' of the temporal graph. These diagrams are powerful tools for analyzing temporal graphs. In Section~\ref{sec:experiments}, we employ these diagrams for classifying temporal graphs.

\subsection{Single Labeled Temporal Graphs}\label{single_filtration}
\subparagraph{\bf{The Average Filtration:}}
Let \( T = (V, E) \) be a single-labeled temporal graph, we construct a filtered simple graph \( G_f = (V_f, E_f, f_\mathrm{avg}: (V_f \cup E_f) \to \mathbb{R}) \) derived from \( T \).  The vertex set \( V_f \) of \( G_f \) is identical to \( V \), and the edge set \( E_f \) of \( G_f \) corresponds to the edges in the aggregate graph \( G \) of \( T \). Specifically, for each temporal edge \( (u, v, t) \in E \) in \( T \), there exists a corresponding edge \( (u, v) \in E_f \) in \( G_f \). Notably, \( G_f \) is a simple graph, meaning that no multiple edges exist between any pair of vertices. The filtration \( f_\mathrm{avg}: (V_f \cup E_f) \to \mathbb{R} \) is referred to as the \textbf{average filtration}.

To motivate the definition of the average filtration \( f_\mathrm{avg} \), we first introduce the concept of the \textbf{minimum filtration} \( f_\mathrm{min}: (V_f \cup E_f) \to \mathbb{R} \). We begin by assigning a filtration value of \( 0 \) to all vertices in \( V_f \), i.e., \( f_\mathrm{avg}(V_f) = 0 \). We then describe the method for computing the filtration values of the edges in \( G_f \).

Let \( \nedges(u,v) := \{ (u', v', t) \mid u = u' \text{ or } v = v' \} \setminus \{ (u, v, t) \} \) represent the set of adjacent temporal edges of \( (u, v, t) \) in \( T \), where each interaction \( (u, v, t) \) belongs to \( T \). Additionally, let \( \tim(u, v) := t \) denote the time label of the edge \( (u, v) \) in \( T \).
%Note that $\tim$ is well defined since $G$ is a single labeled temporal graph.
The filtration value of each edge \( (u,v) \) in \( G_f \) is computed using the minimum of the difference in time stamps between the edge \( (u,v) \) and its adjacent interactions in \( T \):
$$ f_\mathrm{min}(e) := \min_{e' \in \nedges(e)} | \tim(e) - \tim(e')|.$$

As previously mentioned, we aim to capture the evolution of \(\delta\)-temporal motifs within a temporal graph through temporal filtrations. By fixing the values of \(k\) (number of nodes) and \(l\) (number of edges), one can canonically define a filtration over the temporal graph by varying the parameter \(\delta\). Specifically, for each edge \((u, v)\) in a temporal graph, its filtration value is assigned as the smallest \(\delta\) for which it belongs to a fixed $k$ and $l$ \(\delta\)-temporal motif. For \(k = 3\) and \(l = 2\), this canonical filtration corresponds to the minimum filtration \(f_\mathrm{min}\), as defined above. The minimum filtration \(f_\mathrm{min}\) is highly sensitive to changes in interactions, particularly those corresponding to the minimum value, and it often fails to robustly capture the relational and global `temporal connectivity' of the graph. To address this limitation, we redefine the filtration value of an edge as the average of all \(\delta\)-values in the smallest (w.r.t to the parameter $\delta$) \(\delta\)-temporal motifs that include the edge. Specifically, the \textit{average filtration} of the edges in \(G_f\) is defined as:

$$ f_\mathrm{avg}(e) := \frac{\sum_{e' \in \nedges(e)} | \tim(e) - \tim(e')|}{|\nedges(e)|}.  $$

%\sid{Question: How would a temporal graph of one edge would work? We should also discuss what would a maximum filtration do.}
The average filtration reflects the relative temporal importance of an edge: a smaller filtration value indicates that more and faster temporal paths pass through the edge.
 
\begin{figure}[h!]
\begin{minipage}{0.3\textwidth}
\centering
\begin{tikzpicture}[scale = 0.6, -, >=latex, node distance=2cm, thick,scale=0.85]
        % Nodes
        \node (A) at (0, 2) {A};
        \node (B) at (2, 4) {B};
        \node (C) at (4, 2) {C};
        \node (D) at (3, 0) {D};
        \node (E) at (1, 0) {E};
        \node (F) at (6, 4) {F};

        % Temporal edges with timestamps
        \path[every node/.style={font=\sffamily\small}]
            (A) edge node[above left] {$t=1$} (B)
            (B) edge node[above right] {$t=3$} (C)
            (C) edge node[right] {$t=6$} (D)
            (D) edge node[below] {$t=8$} (E)
            (E) edge node[left] {$t=10$} (A)
            (B) edge node[above] {$t=5$} (F)
            (F) edge node[below] {$t=9$} (C);

        % Title
        %\node[below of=F, node distance=2cm] {Temporal Graph};
    \end{tikzpicture}
%\caption{A single-labeled temporal graph.}
\label{fig:single_temporal_graph}
\end{minipage}
\hfill
\begin{minipage}{0.3\textwidth}
    \centering
    % Graph with Average Filtration Labels
    \begin{tikzpicture}[scale = 0.6, -, >=latex, node distance=2cm, thick,scale=1]
        % Nodes
        \node (A) at (0, 2) {A};
        \node (B) at (2, 4) {B};
        \node (C) at (4, 2) {C};
        \node (D) at (3, 0) {D};
        \node (E) at (1, 0) {E};
        \node (F) at (6, 4) {F};

        % Edges with Average Filtration Values
        \path[every node/.style={font=\sffamily\small}]
            (A) edge node[above left] {$5.0$} (B)
            (B) edge node[above right] {$3.67$} (C)
            (C) edge node[right] {$2.67$} (D)
            (D) edge node[below] {$1.67$} (E)
            (E) edge node[left] {$5.5$} (A)
            (B) edge node[above] {$3.33$} (F)
            (F) edge node[below] {$3.5$} (C);

        % Title
        %\node[below of=F, node distance=2cm] {Graph with Average Filtration Labels};
    \end{tikzpicture}
\end{minipage} 
\hfill
\begin{minipage}{0.3\textwidth}
    \centering
    % Graph with Min Filtration Labels
    \begin{tikzpicture}[scale = 0.6,-, >=latex, node distance=2cm, thick, scale=0.85]
        % Nodes
        \node (A) at (0, 2) {A};
        \node (B) at (2, 4) {B};
        \node (C) at (4, 2) {C};
        \node (D) at (3, 0) {D};
        \node (E) at (1, 0) {E};
        \node (F) at (6, 4) {F};

        % Edges with Min Filtration Values
        \path[every node/.style={font=\sffamily\small}]
            (A) edge node[above left] {$2$} (B)
            (B) edge node[above right] {$2$} (C)
            (C) edge node[right] {$2$} (D)
            (D) edge node[below] {$2$} (E)
            (E) edge node[left] {$2$} (A)
            (B) edge node[above] {$2$} (F)
            (F) edge node[below] {$3$} (C);

        % Title
        %\node[below of=F, node distance=2cm] {Graph with Min Filtration Labels};
    \end{tikzpicture}
    \end{minipage}
        \caption{The first figure on the left shows a single-labeled temporal graph, followed by the corresponding average and minimum filtrations derived from it in the next two figures.
}\label{fig:avg_min_filt}
\end{figure}
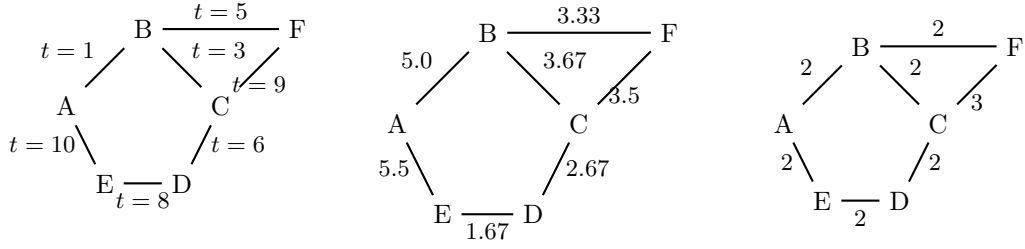

The examples in Figure~\ref{fig:avg_min_filt} illustrate the distinction between the average and minimum filtrations. The temporal loop \(ABCDEA\) in the original temporal graph (left of Figure~\ref{fig:avg_min_filt}) has a temporal length of 9 and can be constructed from smaller \(\delta\)-temporal motifs with 3 nodes and 2 edges for \(\delta = 9\). However, in the minimum filtration, the cycle appears at \(\delta = 2\), while in the average filtration, it emerges at \(\delta = 5.5\). Additionally, the average filtration produces more widely distributed filtration values. This difference suggests that the average filtration takes into account a broader neighborhood around each edge, assigning values that more accurately reflect the relative `temporal position' of the edges. This characteristic makes the average filtration more stable and discriminative compared to the minimum filtration. Experimental results further validate this observation.

We fix the size of the \( \delta \)-temporal motifs to be \( 3 \)-node, \( 2 \)-edge, as it is the smallest (and perhaps only) motif that captures the complete global connectivity of the temporal graph. The example in Figure~\ref{fig:four_motifs} clearly illustrates this; all possible \( 3 \)-node, \( 3 \)-edge \( \delta \)-temporal motifs in the figure would fail to cover the entire temporal graph.

\subparagraph{Algorithm: }
We present an efficient algorithm for computing the average filtration of a temporal graph \( T = (V, E) \). The main idea of the algorithm is as follows: we first iterate over the vertices \( V \) of the graph. For each vertex \( v \in V \), we examine the set of edges \( E_v \subset E \) incident to \( v \). Any two such incident edges \( e \) and \( e' \) will be adjacent and will contribute to each other's filtration values. Specifically, for each pair of incident edges, we compute the timestamp difference \( |t - t'| \), where \( t \) and \( t' \) are the respective timestamps of \( e \) and \( e' \).

To efficiently track these contributions, we maintain a sum variable \( \mathsf{S}_e \) for each edge \( e \). After calculating \( |t - t'| \), we update the sum variables for both edges as follows:
\(
\mathsf{S}_e \gets \mathsf{S}_e + |t - t'| \quad \text{and} \quad \mathsf{S}_{e'} \gets \mathsf{S}_{e'} + |t - t'|.
\)
This process is repeated for all possible pairs of incident edges on \( v \), and the vertex \( v \) is then marked as visited. Once both boundary vertices of an edge \( e = uv \) are marked as visited, we compute its filtration value:
\(
f_\mathrm{avg}(e) = \frac{\mathsf{S}_e}{td_u + td_v},
\)
where \( td_u \) and \( td_v \) are the temporal degrees of the vertices \( u \) and \( v \), respectively. See the pseudocode (Algorithm~\ref{alg:avg_filt}) for more details.

%To optimize the computation of incident temporal edges at each vertex, we store the graph as an adjacency link list of edges incident to each vertex \( v \). This makes the computation of $E_v$ in constant $\mathcal{O}(1)$ time. For each vertex we need to compare $\mathcal{O}(d_{max}^2)$ pair of incident edges, where $d_{max}$ is the maximum temporal degree of the graph. Which results in \( \mathcal{O}(|V| \times d_{max}^2) = \mathcal{O}(|E| \times d_max}\), time complexity.

To optimize the computation of incident temporal edges at each vertex, the graph is stored as an adjacency linked list of edges incident to each vertex \( v \). This representation allows the retrieval of \( E_v \) in constant \( \mathcal{O}(1) \) time. For each vertex, we compare \( \mathcal{O}(d_{\text{max}}^2) \) pairs of incident edges, where \( d_{\text{max}} \) is the maximum temporal degree of the graph. Consequently, the overall time complexity of the algorithm is: \(
\mathcal{O}(|V| \times d_{\text{max}}^2) = \mathcal{O}(|E| \times d_{\text{max}}),
\) where \( |V| \) is the total number of vertices and  \( |E| \) is the total number of temporal edges in the graph.

%To compute the average filtration, we iterate over each edge and calculate its filtration value by traversing its edge neighborhood. The overall time complexity of the algorithm is \( \mathcal{O}(|E| \times d_{max}) \), where \( |E| \) is the total number of temporal edges in the temporal graph \( T = (V, E) \). The following pseudocode provides further details of the implementation.

\begin{algorithm}
\caption{ComputeAverageFiltration }\label{alg:avg_filt}
\begin{algorithmic}[1]
%\Require Temporal graph \( T = (V, E) \), where \( E = \{(u, v, t)\} \)
%\Ensure Filtration values \( f_\mathrm{avg}(e) \) for each edge \( e \in E \)

\State Initialize \( \mathsf{S}_e \gets 0 \) for each \( e \in E \) and \( \mathsf{visited}_v \gets \text{False} \) for each \( v \in V \)
%\State Initialize \( \mathsf{visited}_v \gets \text{False} \) for each \( v \in V \)
\State Initialize an empty filtration value map \( f_\mathrm{avg} \)

\For{each vertex \( v \in V \)}
%    \If{\( \mathsf{visited}_v = \text{True} \)}
%        \State \textbf{continue}
%    \EndIf

    \State Retrieve all incident edges \( E_v = \{e = (v, u, t)\} \)
    \State Initialize a stack \( \mathsf{stack} \) with all edges \( e \in E_v \)
    
    \While{\( \mathsf{stack} \text{ is not empty} \)}
        \State Pop an edge \( e = (v, u, t) \) from \( \mathsf{stack} \)
        \For{each remaining edge \( e' = (v, w, t') \) in \( \mathsf{stack} \)}
            \State Compute \( \Delta \gets |t - t'| \)
            \State Update \( \mathsf{S}_e \gets \mathsf{S}_e + \Delta \)
            \State Update \( \mathsf{S}_{e'} \gets \mathsf{S}_{e'} + \Delta \)
        \EndFor
        
        \If{\( \mathsf{visited}_u = \text{True} \)} \Comment{ $v$ being visited is not required}
            \State Compute \( f_\mathrm{avg}(e) \gets \frac{\mathsf{S}_e}{td_v + td_u} \)
        \EndIf
    \EndWhile
    
    \State Mark \( \mathsf{visited}_v \gets \text{True} \)
\EndFor

\State \Return \( f_\mathrm{avg}(e) \) for all \( e \in E \)

\end{algorithmic}
\end{algorithm}

%Now to compute the average filtration we iterate over each edge and 
%\begin{itemize}
%    \item For each $e_i = (u_i,v_i,t_i) \in T = (V, E)$
%     \begin{enumerate}
%        \item $wt(e_i) = 0$
%        \item $ngh(e_i) = 0$
%        \item For each adjacent edge $e_j$ of $e_i$
%        \begin{enumerate}
%            \item $wt(e_i) = wt(e_i) + |t_j - t_i|$ 
%            \item $ngh(e_i) = ngh(e_i) + 1$ 
%        \end{enumerate}
%        \item if $ngh(e_i) > 0$ then $wt(e_i) = \frac{wt(e_i)}{ngh(e_i}$
%    \end{enumerate}
%    \item EndFor
%\end{itemize}

%\textbf{Complexity :} Should be $O(|E|*k)$, where $|E|$ is the total number of edges and $k$ the degree of the graph.
%
%\textbf{Proof of correctness :} For each edge the smallest time respecting subgraph must contain its neighbours as all time respecting subgraphs are connected. 

\subsection{Multi-labeled Temporal Graphs}
We now extend the average filtration of single-labeled temporal graphs to multi-labeled temporal graphs. %The main challenge arises when the filtered graph becomes non-simple, due to multiple interactions between two nodes at different timestamps. The clique complex of such a graph is not strictly simplicial, complicating the persistence diagram computation. To address this
Our approach for multi-labeled graphs follows a similar methodology as the single-labeled case. In this context, we average the time differences across multiple interactions and assign a single edge to represent the interactions.

Given a multi-labeled temporal graph \( T = (V, E) \), we construct a filtered simple graph \( G_f = (V_f, E_f, f_\mathrm{avg}^\mathrm{mlt}: (V_f \cup E_f) \to \mathbb{R}) \) derived from \( T \). Similar to the single-labeled case, the vertex set \( V_f \) and the edge set \( E_f \) of \( G_f \) correspond to the vertices and edges in the aggregate graph \( G \) of \( T \). Here, \( \tim(u,v) = \{ t \mid (u,v,t) \in E\} \) represents the set of all time labels associated with the edge \( (u,v) \). As in the single-labeled case, the filtration value of vertices is set to 0. The filtration value of the edges in \( G_f \) is computed as follows:

$$
f_\mathrm{avg}^\mathrm{mlt}(e) = \frac{\sum_{e' \in \nedges(e)} \sum_{t \in \tim(e), t' \in \tim(e')} | t - t'|}{|\nedges(e)|}.
$$

%\sid{Why not average over all the interactions rather than the number of  edges (pair of nodes).}

%Here too the filtration value of an edge corresponds to the average of all \(\delta\)-values of the smallest (w.r.t to the parameter $\delta$) \(\delta\)-temporal motifs that include the edge. 

%In the second approach, we compute the average of the time labels for interactions between a pair of nodes and assign this average as a single time label. This converts the multi-labeled temporal graph \( T \) into a single-labeled temporal graph \( T_s \). The vertices of \( T_s \) are the same as those in \( T \), and for each edge \( e \), we define the time label in \( T_s \) as:

%\[\tim_{s}(e) = \frac{\sum_{t \in \tim(e)} t}{|\tim(e)|}.\]

%Here, \( \tim_{s}(e) \) is the unique time stamp for each temporal edge in \( T_s \). The single-labeled graph \( T_s \) is then used to compute the average filtration.

%Now we convert $G_S$ to a weighted graph as described in Section~\ref{single_filtration} to compute the filtration.
%
%
%Once we create the persistence diagrams for each graph, they are stored in a text file separated by newlines. 
%The file is now parsed through a C++ program that calculates the Persistence Scale Space Kernel (cite reninghaus), which is stored in a csv.
%
%The kernel matrix is then split into train and test sub matrices according to the graphs split in train and test sets previously. 
%The precomputed kernel matrix for the train set is now fitted to a vanilla SVM classifier, and then the accuracy is tested on the test submatrix.

\section{Stability}\label{sec:Stability}

In this section, we examine the stability of the temporal filtrations defined earlier in the context of \textit{randomized reference models} of temporal graphs. To provide a foundation, we first briefly review the concept of randomized reference models and introduce some commonly used models, as outlined in \cite{Holme2012, Karsai2011}. These models form the basis for defining the various classification classes used in our experiments.

\subsection{Randomized Reference Model}
The \textit{configuration model} is a randomized reference model, commonly used in the study of static networks to compare the empirical network's features (such as clustering, path length, or other topological characteristics) with those of a randomized network. The configuration model is created by randomly shuffling the edges of a given graph while preserving some of its structural properties, most notably the degree distribution.

%For static networks, the significance or anomaly of topological features is often assessed by comparing them to a randomized reference model, such as the configuration model, which preserves the original degree sequence while randomizing links. This allows for evaluating the significance of graph characteristics through direct comparisons, statistical test scores, or differences in process dynamics between the empirical network and the reference

For temporal graphs, a similar approach involves randomizing or reshuffling event sequences (interactions) to remove time-domain structures and correlations. Unlike static networks, temporal graphs exhibit diverse temporal correlations (structures) across varying scales, making it challenging to design a single, universal null model. Karsai et.al.~\cite{Karsai2011} identify five of these temporal correlations, namely: community structure (C), weight-topology correlations (W), bursty event dynamics on single links (B), and event-event correlations between links (E) and a daily pattern (D). They also provide tailored null models that selectively remove specific correlations to analyze their influence on the observed temporal features or dynamical processes like spreading. %By comparing dynamics across models, the role of different temporal and topological correlations can be assessed.
Below, we recall three temporal null models from Karsai et al.~\cite{Karsai2011} and Holme~\cite{Holme2012}. 
 
\begin{itemize}
    \item \textbf{Equal-Weight Link-Sequence Shuffle (EWLSS):}  
    In this method, two interaction pairs \((u_1, v_1), (u_2, v_2) \in V \times V\) are randomly selected such that \(|\tim(u_1, v_1)| = |\tim(u_2, v_2)|\). Then the time labels of these selected pairs \((u_1, v_1)\) and \((u_2, v_2)\) are swapped. Note that, \(\tim(u, v)\) represents the set of time labels associated with the pair \((u, v)\). This process can be repeated multiple times to construct a new randomized temporal graph. %This process erases the previous temporal correlations between edges but retains the event counts. %For links with large weights, events are grouped into bins with 2–3 weight values to simplify the shuffling process.
	\item \textbf{Randomized Edges (RE):} Algorithmically this method could be described as follows: iterate over all edges and for each edge $(u,v)$, select another edge $(u', v')$.  With 50\% probability, replace $(u,v)$ and $(u', v')$ with $(u, v')$ and $(u', v)$; otherwise, replace them with $(u, u')$ and $(v, v')$. The times of contact for each edge remain constant and we further make sure that there are no self loops or multiple edges. %This method extends the configuration model for static graphs by preserving edge-specific contact sequences during rewiring.
 %\item \textbf{DCB (Link-Sequence Shuffled):}  Whole single-link event sequences are randomly exchanged between randomly chosen links, without considering event counts. This method eliminates both event-event correlations and weight-topology correlations.

    %\item \textbf{DCW (Time-Shuffled):}  The time stamps of all events in the original sequence are shuffled randomly. This destroys temporal correlations while preserving the overall distribution of events.

    \item \textbf{Configuration Model (CM):}  
    The aggregated graph of the temporal graph is rewired using the configuration model of the static graph, which preserves the degree distribution and overall connectivity of nodes while removing topological correlations. Then the original single-edge interaction time labels are randomly assigned to the edge, followed by time shuffling. This method destroys all correlations except broad seasonal patterns, such as daily cycles.
\end{itemize}

In addition to these three model we introduce a new null model that is suitable for some of our specific experiments.

\begin{itemize}
\item \textbf{Time Perturbation (TP):} In this method, a fraction of interactions \( e = (u, v, t) \in E \) in the original temporal graph is replaced with \( e' = (u, v, t') \), where \( |t - t'| < \epsilon \) for some \( \epsilon > 0 \). This procedure only perturbs the time stamps of the edges, preserving most of the temporal and structural features of the original temporal graph.
\end{itemize}

%This null model isolates the effect of network topology while preserving temporal correlations and edge-specific patterns like `burstiness' and inter-contact time distributions. It assumes edges govern contact times, meaning vertex contact timings and counts may vary post-randomization, but vertex degrees in the aggregated network remain unchanged. Additionally, the overall event rate over time is preserved. An illustration of this process can be found in Fig.~12.

The first model, EWLSS, preserves the daily pattern (D), the community structure (C), weight-topology correlations (W), and bursty event dynamics on individual edges (B). In contrast, the configuration model loses all temporal correlations (structures) except for the daily pattern (D). This distinction has been experimentally verified by Karsai et al.~\cite{Karsai2011}, where graphs generated using EWLSS procedures exhibit spreading dynamics closely resembling those of the original graph. On the other hand, graphs generated using the configuration model deviate significantly from the original dynamics.

We leverage these models to define and populate distinct classes for classification tasks. The configuration model generates temporal graphs with diverse dynamics, while the other three models (EWLSS, RE and TP) are used to produce graphs with similar dynamics, forming a single class. The similarity within each class depends on the model used; TP and EWLSS create more homogeneous classes compared to the RE model. Originally designed for studying spreading dynamics with the SI model~\cite{kermack1927mathematical}, these null models eliminate the need for direct SI simulations, enhancing both the efficiency and flexibility of our method without requiring labeled nodes.

\subsection{Stability} 
We now discuss the stability of our temporal filtration with respect to the above null models. In particular, we calculate the difference in the average filtrations values of the edges that are , shuffled, swapped or changed during each step of the above reference modes.

\subparagraph*{{TP Procedure :}}

%Given a temporal graph $T = (V, E)$, a new temporal graph is constructed by randomly perturbing a percentage of the interactions in $G$.  That is, for some $\epsilon > 0$, we replace of percentage of interactions $e = (u,v,t) \in E$ with $e' = (u , v, t')$ such that $| t - t'| < \epsilon $. This procedure only perturbs the time stamps of the edges and preserves the structure of the temporal graph. 

Let $T$ be a single-labeled temporal graph and $e = (u,v,t)$ is an interaction in $T$. Suppose the time label \( \tim(e) = t \) is replaced by \( t + \epsilon \).  The change in \( f_\mathrm{avg}(e) \), denoted by \( \Delta f_\mathrm{avg}(e) \), can be upper bounded as follows:

%Let \( f_\mathrm{avg}(e) \) denote the single-label average filtration of an edge \( e \), given by
%\[f_\mathrm{avg}(e) := \frac{\sum_{e' \in \nedges(e)} | \tim(e) - \tim(e')|}{|\nedges(e)|}.\]

\begin{align*}
    \Delta f_\mathrm{avg}(e) & = f_\mathrm{avg}(e)_{\text{new}} - f_\mathrm{avg}(e)_{\text{old}} \\ 
    & = \frac{1}{|\nedges(e)|} \sum_{e' \in \nedges(e)} \Big( |(t + \epsilon) - \tim(e')| - |t - \tim(e')| \Big).
\end{align*}

% $$
% \Delta f_\mathrm{avg}(e) = f_\mathrm{avg}(e)_{\text{new}} - f_\mathrm{avg}(e)_{\text{old}} = \frac{1}{|\nedges(e)|} \sum_{e' \in \nedges(e)} \Big( |(t + \epsilon) - \tim(e')| - |t - \tim(e')| \Big).
% $$

Let \( \tim(e') = t' \), then for each \( e' \in \nedges(e) \) the inner term is
$
| (t + \epsilon) - t'| - |t - t'| \leq \epsilon.
$
%This follows from the triangle inequality, which ensures that the change in absolute difference is bounded by \( |\epsilon| \), the shift in \( t \). Similarly:
%\[
%| (t + \epsilon) - t'| - |t - t'| \geq -\epsilon.
%\]
%
%Thus, the term \( |(t + \epsilon) - t'| - |t - t'| \) satisfies:
%\[
%-\epsilon \leq |(t + \epsilon) - t'| - |t - t'| \leq \epsilon.
%\]
%
%Summing Over Edges
%The sum \( \sum_{e' \in \nedges(e)} \) aggregates contributions from all edges \( e' \in \nedges(e) \). With \( |\nedges(e)| \) total edges, the cumulative change is bounded by:
%\[
%\sum_{e' \in \nedges(e)} \Big( |(t + \epsilon) - \tim(e')| - |t - \tim(e')| \Big) \leq \epsilon \cdot |\nedges(e)|.
%\]
%
% Final Bound
%Dividing by \( |\nedges(e)| \) to compute the average, 
we therefore obtain:
$
|\Delta f_\mathrm{avg}(e)| \leq \epsilon.
$ The filtration values of edges in the neighborhood of \( e \) also change. Using a similar computation and the fact that $\tim(e'')$ does not change for all $e'' \in \nedges(e') \setminus \{e\}$, for each \( e' \in \nedges(e) \), we can express the change as:

\begin{align*}
    \Delta f_\mathrm{avg}(e') & 
= \frac{1}{|\nedges(e')|} (|(t + \epsilon) -  \tim(e')| - |t -  \tim(e')|) = \frac{\epsilon}{|\nedges(e')|} \leq \epsilon.
\end{align*}

% Hence, 
% \(
% |\Delta f_\mathrm{avg}(e')| = \frac{\epsilon}{|\nedges(e')|} \leq \frac{\epsilon}{d_{max}},
% \)
% where \( d_{max} > 1 \) represents the maximum degree of the graph \( T \).

We use the \( L_{\infty} \) norm to measure the distance between filtrations~\cite{cohen2007stability}. The overall distance between the new (shifted) and the original average filtration of the temporal graph \( G \) is bounded above as follows:
\(
|\Delta_{\infty} f_\mathrm{avg}(G)| \leq \epsilon.
\)
Note that this bound holds even after time stamp shifts of multiple edges.

%$$  | f_\mathrm{avg}(e) - f'_{avg} (e') | = $$ 
%$$ = \frac{| \sum_{e^n \in \nedges(e)} \sum_{t_G \in \tim(e) , t^n_G \in \tim(e^n)}| t_G - t^n_G| - \sum_{e^n \in \nedges(e')} \sum_{t'_{G'} \in \tim'(e') , t^n_{G'} \in \tim'(e^n)}| t'_{G'} - t^n_{G'}| |}{|\nedges(e)|} $$
%$$ =  \frac{| \sum_{e^n \in \nedges(e)} \sum_{t_G \in \tim(e) , t'_G \in \tim(e')}| t_G - t'_G| -  \sum_{t_{G'} \in \tim' (e) , t'_{G'} \in \tim'(e')}| t_{G'} - t'_{G'}| |}{|\nedges(e)|}$$
%$$ \leq \frac{\sum_{e' \in \nedges(e)} 2 \epsilon }{|\nedges(e)|} = 2 \epsilon $$

For simplicity, we have assumed that \( T \) is a single-labeled temporal graph. However, through a similar calculation, it can be shown that the same bound applies to multi-labeled temporal graphs and cases where multiple time labels of an edge are shifted. Since the underlying aggregate graph remains unchanged after a TP procedure, the stability theorem~\cite{cohen2007stability} guarantees stability in the resulting persistence diagrams.

%This gives us the stability property for the TP Procedure.

\subparagraph{\textbf{EWLSS Procedure:}} To upper bound the differences in the average filtration when swapping time labels \(t_1\) and \(t_2\) for the edges \(e_1 = (u_1, v_1, t_1)\) and \(e_2 = (u_2, v_2, t_2)\), we proceed as follows:

\[ \text{For } e_1:
\Delta f_\mathrm{avg}(e_1) = \frac{\sum_{e' \in \nedges(e_1)} \Big(|t_2 - \tim(e')| - |t_1 - \tim(e')|\Big)}{|\nedges(e_1)|}.
\]
Again we can bound the inner term, 
$
\Big| |t_2 - \tim(e')| - |t_1 - \tim(e')| \Big| \leq |t_2 - t_1|,
$ Thus:
$$
|\Delta f_\mathrm{avg}(e_1)| \leq \frac{|\nedges(e_1)| \cdot |t_2 - t_1|}{|\nedges(e_1)|} = |t_2 - t_1|.
$$

By symmetry, $|\Delta f_\mathrm{avg}(e_2)| \leq |t_2 - t_1|.$ The filtration values of the edges in the neighborhoods \(\nedges(e_1)\) and \(\nedges(e_2)\) of \(e_1\) and \(e_2\), respectively, will also change. However, these changes will be bounded above by $|t_2 - t_1|$.
%\(\frac{|t_2 - t_1|}{d_{\max}}\), where \(d_{\max}\) is the maximum degree of the graph.

For a single pair swap, the absolute distance between the average filtration of the swapped graph and the original graph is bounded as:
\(
|\Delta_{\infty} f_\mathrm{avg}(G)| \leq |t_2 - t_1|.
\)
For multiple pair swaps, the distance is bounded above by the maximum time label difference among the pairs:
\(
|\Delta_{\infty} f_\mathrm{avg}(G)| \leq \max{|t_2 - t_1|}.
\)
As in the previous case, the underlying aggregate graph remains unchanged after an EWLSS procedure. Consequently, the persistence diagrams remain stable as well.

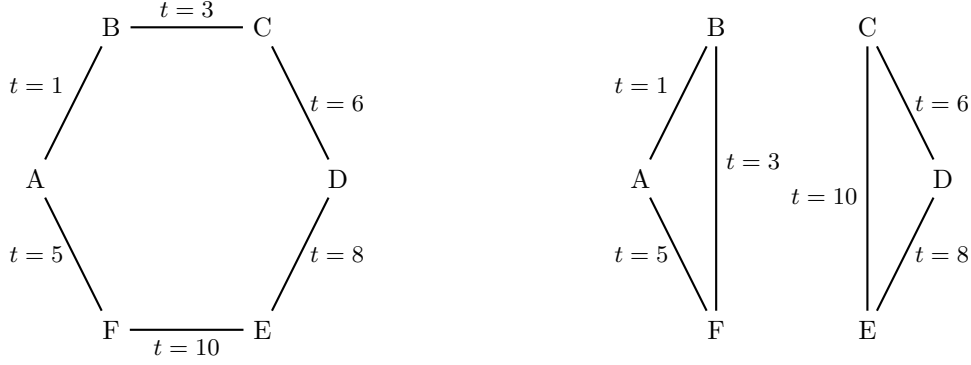
\begin{figure}[hpt]
    \centering
    \begin{tikzpicture}[-, >=latex, node distance=2cm, thick, scale = 1]
        % Nodes (Left Graph)
        \node (A) at (-5, 2) {A};
        \node (B) at (-4, 4) {B};
        \node (C) at (-2, 4) {C};
        \node (D) at (-1, 2) {D};
        \node (E) at (-2, 0) {E};
        \node (F) at (-4, 0) {F};

        % Edges before RE
        \path[every node/.style={font=\sffamily\small}]
        (A) edge node[above left] {$t=1$} (B)
        (B) edge node[above] {$t=3$} (C)
        (C) edge node[right] {$t=6$} (D)
        (D) edge node[right] {$t=8$} (E)
        (E) edge node[below] {$t=10$} (F)
        (F) edge node[left] {$t=5$} (A);

        % Nodes (Right Graph)
        \node (A') at (3, 2) {A};
        \node (B') at (4, 4) {B};
        \node (C') at (6, 4) {C};
        \node (D') at (7, 2) {D};
        \node (E') at (6, 0) {E};
        \node (F') at (4, 0) {F};

        % Edges after RE
        \path[every node/.style={font=\sffamily\small}]
        (A') edge node[above left] {$t=1$} (B')
        (B') edge node[above right] {$t=3$} (F')
        (C') edge node[right] {$t=6$} (D')
        (D') edge node[right] {$t=8$} (E')
        (E') edge node[below left] {$t=10$} (C')
        (F') edge node[left] {$t=5$} (A');
    \end{tikzpicture}
    \caption{An example of a single step in the RE procedure involves shuffling the edges \( BC \) and \( FE \) to \( BF \) and \( CE \), respectively, while maintaining their original time stamps.} \label{fig:re_stability}
\end{figure}

\subparagraph{\textbf{RE and CM Procedure:}} The average filtration does not exhibit theoretical stability for the remaining two procedures, RE and configuration models. This behavior is expected, primarily because the underlying aggregate graphs could change at each step of these procedures. Such changes result in an infinite difference between the filtration values of previously non-existent interactions and those of newly added interactions or vice-versa. We illustrate this change for the RE procedure in Figure~\ref{fig:re_stability}. Among the RE and configuration models, RE generates more similar temporal graphs. Therefore, to design a relatively heterogeneous class, we still use the RE procedure to populate a single class.

\section{Experiments}\label{sec:experiments}
In this section, we present experiments to evaluate the effectiveness of our method in classifying temporal graphs. 
\subparagraph{\textbf{Pipelines:}}
Our benchmarking of temporal graph classification uses two pipelines:
\begin{enumerate}[\label=\alph*)]

    \item \textbf{Persistent Homology pipeline:}
    For each temporal graph, we compute the average filtration and its
    corresponding flag filtration, from which we extract persistence
    diagrams up to degree~2. For each degree, we compute a kernel
    distance matrix using the Persistence Scale Space
    kernel~\cite{reininghaus2015stable}, yielding three matrices (degrees
    0--2). These are combined with equal weights\footnote{Different
    weights allow adjusting the influence of each degree.} to form a
    single kernel matrix, which is then used to train an SVM classifier.

    \item \textbf{Graph filtration kernel pipeline:}
    For each temporal graph, we compute the average filtration along with
    the static graph filtrations of~\cite{Schulz_2022} for benchmarking. We then apply the
    filtered Weisfeiler--Lehman kernel~\cite{Schulz_2022} to obtain a
    kernel matrix, which is used to train an SVM classifier. Except for
    computing the average filtration, all subsequent steps use the
    implementation provided in the
    \href{https://github.com/mlai-bonn/wl-filtration-kernel}{source code}
    of~\cite{Schulz_2022}.
\end{enumerate}

\subparagraph{Class Population:}
To train the SVM classifier, we require at least two classes of temporal
graphs. We follow a method similar to~\cite{DallAmico2024} to generate
and populate these classes. Real datasets typically provide only a single
temporal graph (the \textbf{root graph}), whereas synthetic datasets may
contain one or more root graphs. The TP, EWLSS, and RE procedures
(Section~\ref{sec:Stability}) generate loosely similar copies of a root
graph, while the CM procedure produces distinct ones. This distinction is
experimentally verified in~\cite{Karsai2011} through SI-model simulations
and analysis of infected-node distributions. For real datasets with a
single root graph, two classes are formed: one of loose copies and one of
CM-generated graphs\footnote{There are (at least) two possible ways to generate multiple classes.
In the first, a CM-generated variant of the root graph serves as the
representative of the second class, and both classes are populated using
RE, TP, or EWLSS. In the second, the root graph represents the first
class and is populated by loose copies, while the second class is formed
by multiple distinct CM-generated variants. Thus, the first class
contains similar graphs, whereas the second contains structurally
different ones.}. With multiple root graphs, each root graph yields its own class through its loose copies.

\subparagraph{\bf{Datasets}:} We summarize below the four real contact-sequence datasets used to
benchmark our method. All datasets are sourced from the SocioPatterns
repository~\cite{sociopatterns_datasets}. 

%The datasets used in the first two experiments are contact sequence datasets. The first dataset is a real-world dataset from the SocioPatterns dataset repository~\cite{sociopatterns_datasets}. %We have used the Hospital Ward~\cite{hospital_data} and High-School~\cite{highschool} dynamic contact network datasets, Workplace contacts~\cite{workplacev2} dataset and a temporal network of student contacts at MIT~\cite{Eagle2006}. 

\begin{table}[ht]
    \centering
    %\resizebox{1\textwidth}{!}{
    \begin{tabular}{|l|l|l|l|l|l|l|l|}
        \hline
        \textbf{Dataset} & \textbf{$|V|$} & \textbf{$|E_s|$} & \textbf{$|E|$} & \textbf{$E_{avg}$} & \textbf{$E_{max}$} & \textbf{$d_{avg}$} & \textbf{$d_{max}$} \\ \hline
        Highschool 2011 & 126 & 1710 & 28540 & 16.69 & 1185 & 27.14 & 55 \\ \hline
        Highschool 2012 & 180 & 2220 & 45047 & 20.29 & 1280 & 24.67 & 56 \\ \hline
        Hospital Combined & 75 & 1139 & 32424 & 28.47 & 1059 & 30.37 & 61 \\ \hline
        %MIT & 96 & 2539 & 234757 & 92.46 & 4387  & 52.9 & 92 \\ \hline
        Workplace v2 & 217 & 4274 & 78249 & 18.31 & 1302 & 39.39 & 84 \\ \hline
    \end{tabular} %}
    \caption{$|V|$ denotes the number of vertices, $|E_s|$ the number of edges in the
aggregate graph, and $|E|$ the number of temporal edges. $E_{\text{avg}}$
and $E_{\text{max}}$ are the average and maximum number of time labels
per edge, while $d_{\text{avg}}$ and $d_{\text{max}}$ are the average and
maximum temporal degrees of the nodes.
}
\label{tab:dataset_stats}
\end{table}

\begin{figure}[hpt]
\centering
\begin{tikzpicture}
\begin{axis}[
    ybar,
    bar width=3pt,
    width=12cm,
    height=7cm,
    ymin=0.5, ymax=1.01,
    enlarge x limits=0.15,
    symbolic x coords={
        Hospital Combined,
        Hospital (W/NW),
        Workplace v2,
        HighSchool 2011,
        HighSchool 2012,
        HighSchool (2011/2012)
    },
    xtick=data,
    xticklabel style={rotate=45, anchor=east},
    ylabel={Accuracy},
    legend style={at={(0.5,1.05)}, anchor=south, legend columns=-1},
]

% PH pipeline
\addplot coordinates {
    (Hospital Combined,0.98625)
    (Hospital (W/NW),0.925)
    (Workplace v2,1.0)
    (HighSchool 2011,1.0)
    (HighSchool 2012,1.0)
    (HighSchool (2011/2012),0.9866)
};

% Filtered-kernel pipeline
\addplot coordinates {
    (Hospital Combined,0.9875)
    (Hospital (W/NW),1.0)
    (Workplace v2,1.0)
    (HighSchool 2011,1.0)
    (HighSchool 2012,1.0)
    (HighSchool (2011/2012),1.0)
};

%Static Graph-add_max_degree_weights
\addplot coordinates {
    (Hospital Combined, 0.5875)
    (Hospital (W/NW),  1)
    (Workplace v2,0.8125)
    (HighSchool 2011,0.7475)
    (HighSchool 2012,0.9600)
    (HighSchool (2011/2012),1.0)
};
%Static add_core_num_weight
\addplot coordinates {
    (Hospital Combined, 0.6950)
    (Hospital (W/NW),  1)
    (Workplace v2, 0.9425)
    (HighSchool 2011, 0.8850)
    (HighSchool 2012, 0.7675)
    (HighSchool (2011/2012), 1.0)
};

%\addplot coordinates {
%    (Hospital Combined, 0.6950)
%    (Hospital (W/NW) , 1.0)
%    (Workplace v2, 0.9425)
%    (HighSchool 2011, 0.8850)
%    (HighSchool 2012, 0.7675)
%    (HighSchool (2011/2012),1.0)
%};

%Static add_triangle_weight
\addplot coordinates {
    (Hospital Combined,0.5725)
    (Hospital (W/NW), 1)
    (Workplace v2, 0.9800)
    (HighSchool 2011,0.9650)
    (HighSchool 2012,0.9900)
    (HighSchool (2011/2012),1.0)
};

\legend{PH pipeline, Filtered-kernel pipeline,add-max-deg,add-core-num,add-triangle}
\end{axis}
\end{tikzpicture}
\caption{Comparison of the PH and filtered-kernel pipelines across datasets. The
reported accuracy scores are averaged over multiple runs, with only
marginal variation across trials.}
\label{fig:ph-vs-fk}
\end{figure}
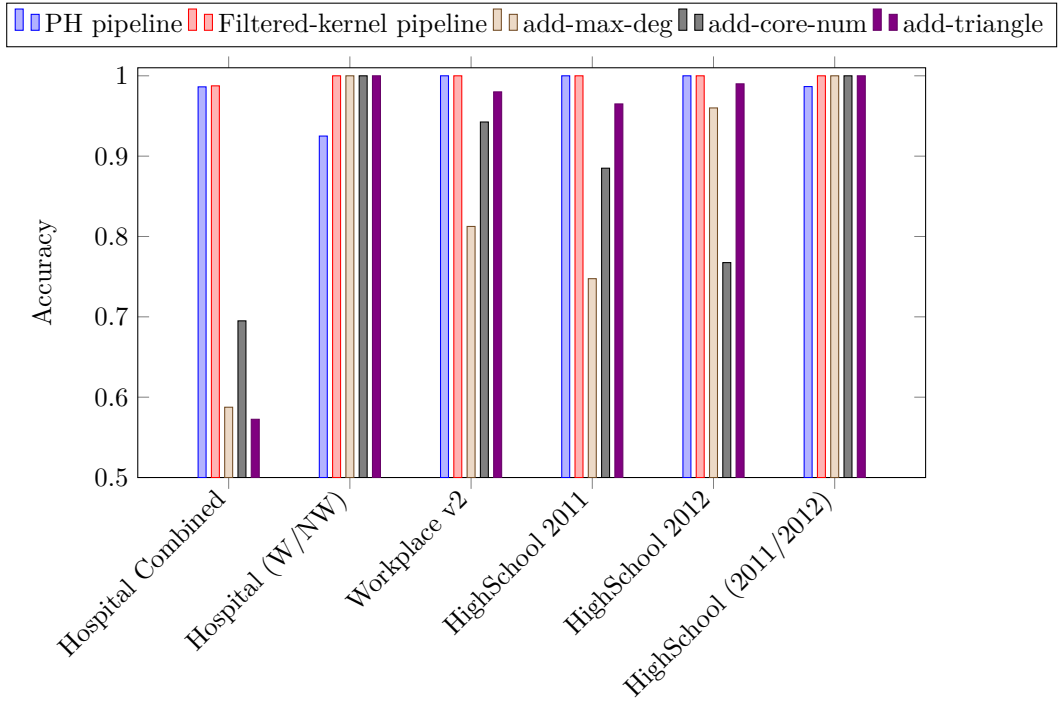

\subparagraph{Benchmark Accuracies:} The histogram in Figure~\ref{fig:ph-vs-fk} shows the classification
accuracies from our experiments. The two temporal-graph classes were
generated as described in the \emph{Class population} paragraph above.
Each dataset was split into training and testing sets using an 80--20
ratio, and accuracy was computed on the testing set as the proportion of
correctly classified instances. The source code for all experiments using PH-pipeline is
publicly available on GitHub at the following
\href{https://github.com/phtgraph/temporal_classification?tab=readme-ov-file}{link}.

Both the PH pipeline and the filtered-kernel pipeline use the average
filtration\footnote{All temporal graphs in our experiments (except the random temporal graph
experiment) are multi-labeled, and we use the
\(f_\mathrm{avg}^\mathrm{mlt}\) filtration for both pipelines, as
described in Section~\ref{sec:temp_filtration}.}
for classification. In addition, the filtered-kernel pipeline incorporates
three static-graph filtrations from~\cite{Schulz_2022}:
\texttt{add-max-deg}, \texttt{add-core-num}, and \texttt{add-triangle} \footnote{%Given a pair of vertices representing an edge, 
The \texttt{add-max-deg} filtration assigns to each edge the maximum
degree of its two endpoints in the aggregate graph. The \texttt{add-core-num}
filtration uses the core numbers of the endpoints, where the core number
of a node is the largest \(k\) such that it belongs to a \(k\)-core
(subgraph in which all nodes have degree at least \(k\)); an edge is
assigned the maximum core number of its endpoints. The \texttt{add-triangle}
filtration assigns to each edge half of the maximum number of triangles
incident to either endpoint.
}.
The average filtration consistently provides strong classification
performance, while the static filtrations also perform well when the
aggregate graph structures of the classes differ sufficiently. For example, the
Hospital (W/NW) dataset consists of two classes obtained from distinct
segments of the Hospital Combined data—the working (staff) and
non-working (patient) segments—and their structural differences lead to
high accuracy for the static kernels. A similar trend is observed in the
HighSchool (2011/2012) experiments.

\subparagraph{Similar Classes.}
We also benchmarked the PH and filtered-kernel pipelines using the average filtration in settings
where the two classes are highly similar—that is, the second-class
representative is generated as a loose copy of the root graph using
additional RE or EWLS steps\footnote{Extra RE/EWLS steps were used to
generate the second-class representative.}. In such cases, the temporal
graphs are extremely similar, making the classes inherently difficult to
distinguish. This is reflected in our experimental results
(Table~\ref{tab:combined_average_real_contact_same_class}), further
supporting the use of reference models to induce meaningful class
differences.

% Nice number alignment:
\sisetup{
    table-number-alignment = center,
    round-mode = places,
    round-precision = 4
}

\begin{table}[ht]
\centering
\begin{tabular}{
    l
    l
    S[table-format=1.4]
    S[table-format=1.4]
    S[table-format=1.4]
    S[table-format=1.4]
}
\toprule
\textbf{Dataset} & \textbf{Classes} 
& \multicolumn{2}{c}{\cellcolor{gray!10}\textbf{PH}} 
& \multicolumn{2}{c}{\cellcolor{gray!10}\textbf{Filtered Kernel}} \\
\cmidrule(lr){3-4}\cmidrule(lr){5-6}
& & \textbf{Acc.} & \textbf{Std.} & \textbf{Acc.} & \textbf{Std.} \\
\midrule

Hospital Combined & RE + RE      
& 0.5075  & 0.08419 
& 0.5625  & 0.07090 \\

Hospital Combined & EWLS + EWLS  
& 0.5325  & 0.06934 
& 0.5275  & 0.04250 \\

HighSchool 2011   & RE + RE      
& 0.45875 & 0.05338 
& 0.4450  & 0.06200 \\

HighSchool 2011   & EWLS + EWLS  
& 0.505   & 0.07068 
& 0.4850  & 0.07840 \\

HighSchool 2012   & RE + RE      
& 0.50625 & 0.05741 
& 0.5375  & 0.04910 \\

HighSchool 2012   & EWLS + EWLS  
& 0.49625 & 0.05016 
& 0.4800  & 0.06200 \\

\bottomrule
\end{tabular}
\caption{In the second column, “RE+RE’’ indicates that both classes are generated using the RE procedure on the root graph; other entries follow the same convention.}
\label{tab:combined_average_real_contact_same_class}
\end{table}

\subparagraph*{{Synthetic Datasets:}} Next, we evaluate the PH pipeline on synthetic contact-sequence datasets
(see Table~\ref{tab:combined_averages_synthetic_contact}). We use three
root graphs with varying parameters, generated under both
\textit{disassortative} and \textit{assortative} mixing
strategies~\cite{Holme2012}, modeling contact and disease-transmission
dynamics.
 %The last two experiments use the same graphs: the third uses the full persistence diagram, while the fourth uses a pruned version where low-persistence points are removed. We observe only a slight drop in accuracy, accompanied by nearly a twofold reduction in runtime. %The last two experiments use the same set of graphs: the third uses all points in the persistence diagram, while the fourth uses a pruned diagram in which low-persistence points are removed. We observe only a minimal drop in accuracy, while the runtime decreases by nearly half.

\begin{table}[h!]
\centering
%\resizebox{1\textwidth}{!}{
\begin{tabular}{|l||c|c|c|c|}
\hline
\textbf{Class} & \textbf{|V|/|E|} & \textbf{Threshold}   & \textbf{Average Accuracy} & \textbf{Average Stdev} \\ \hline
3 & 100/200               & 0   & 0.9733                    & 0.0235 \\ \hline
3 & 250/500               & 0   & 1.0000                   & 0.0000  \\ \hline
3 & 100/1000              & 0   & 0.9681 %(104s)
&  0.0283   \\ \hline
%3 & 100/1000              & 300 & 0.9385 (49s)				  & 0.0396     \\ \hline
\end{tabular} %}
\caption{ %In the first two experiments, we use 18 distinct parameter sets, while in the last one, we use 6. For each parameter set, 
Results are averaged over 5 runs. On average, generating a new class
member requires 20 RE steps in the first two experiments and 32.5 steps
in the last one.  %The runtime for the third and fourth experiments is 104 seconds and 49 seconds, respectively.
 %The mixing functions take 2 parameters to decide if 2 nodes become a pair, $x_i$ which sets the strength of assortativity or disassortativity and $d_{max}$ is the maximum degree a node can take in a graph. 
}
\label{tab:combined_averages_synthetic_contact}
\end{table}

\subparagraph{Random Temporal Graphs:} 
%In Figures~\ref{fig:random_plot1} and~\ref{fig:random_plot2}, we present the results of our experiments on random temporal graphs. Both experiments use single-labeled graphs and the $f_\mathrm{avg}$ filtration, as described in Section~\ref{sec:temp_filtration}. The goal is to assess the effectiveness of our approach with a large number of similar classes and to evaluate its performance under varying temporal graph sparsity. To generate different classes, we apply the TP procedure from Section~\ref{sec:Stability} to a fraction of interactions, referred to as \textit{out shifts}. For a single class, the TP procedure is applied to a smaller fraction of edges, referred to as \textit{in shifts}. More details of this strategy are provided for each experiment.
Figures~\ref{fig:random_plot1} and~\ref{fig:random_plot2} present the
results of experiments on random temporal graphs using single-labeled
graphs and the PH pipeline. These graphs do not model any specific phenomena. These experiments assess our method’s
performance with many similar or mixed classes, as well as under varying
temporal sparsity. For class generation, we use the TP procedure from
Section~\ref{sec:Stability}: \textit{out shifts} apply TP to a fraction
of interactions to create distinct classes, while \textit{in shifts}
apply TP to a smaller fraction of edges within a single class. Details
for each experiment are given in the figure descriptions.

\begin{enumerate}[\label=\alph*)]

\item {\textbf{Pure Classes:}} To create a homogeneous pool of classes, we generated a single-labeled temporal graph $G$ with 100 vertices and varying sparsity (0.05 to 0.8), with time stamps in the range $(0,100]$. For each experiment, the original root graph was used to create 3, 5, 7, or 9 new root graphs (classes) by shifting the time stamps (out-shifts) of an average of 4.75\% of interactions by $1 \leq \epsilon \leq 5$. To populate graphs within the same class, time stamps of an average of 1.6\% of interactions were shifted (in-shifts) by $1 \leq \epsilon \leq 5$.

% A single labeled temporal graph $G$ with 100 vertices and \sid{$e$ how many?} interactions was randomly generated.    Each edge had weights in the range $(0,100]$.    We then create $n$ root graphs $G_1,\dots,G_n$ using TP Procedure.   For each root graph, a class is created using the TP procedure, but on a smaller percentage of intractions.
    
%We provide accuracy results for temporal graphs across various sparcity in 
%Tables~\ref{tab:exp1-0.05},~\ref{tab:exp1-0.1},~\ref{tab:exp1-0.2},~\ref{tab:exp1-0.4}
%and ~\ref{tab:exp1-0.8}.

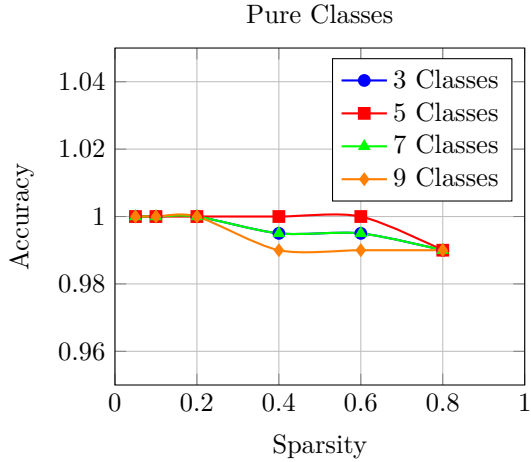
\begin{figure}[h!]
\centering
\begin{minipage}{0.5\textwidth} % Left minipage for the plot
    \begin{tikzpicture}
    \begin{axis}[
            width=1.0\textwidth,
            title={Pure Classes},
            xlabel={Sparsity},
            ylabel={Accuracy},
            xmin=0, xmax=1,
            ymin=0.95, ymax=1.05,
            legend pos=north east,
            grid=both,
            every axis plot/.append style={thick},
            cycle list={
                {blue,mark=*},
                {red,mark=square*},
                {green,mark=triangle*},
                {orange,mark=diamond*}
            }
        ]
    \legend{3 Classes , 5 Classes, 7 Classes, 9 Classes}    
    % Data points for Class 3
    \addplot[
        color=blue,
        mark=*,
        mark options={fill=blue},
        smooth
    ] coordinates {
        (0.05, 1.00)
        (0.1, 1.00)
        (0.2, 1.00)
        (0.4, 0.995)
        (0.6, 0.995)
        (0.8, 0.99)
    };

    % Data points for Class 5
    \addplot[
        color=red,
        mark=square*,
        mark options={fill=red},
        smooth
    ] coordinates {
        (0.05, 1.00)
        (0.1, 1.00)
        (0.2, 1.00)
        (0.4, 1.00)
        (0.6, 1.00)
        (0.8, 0.99)
    };

    % Data points for Class 7
    \addplot[
        color=green,
        mark=triangle*,
        mark options={fill=green},
        smooth
    ] coordinates {
        (0.05, 1.00)
        (0.1, 1.00)
        (0.2, 1.00)
        (0.4, 0.995)
        (0.6, 0.995)
        (0.8, 0.99)
    };

    % Data points for Class 9
    \addplot[
        color=orange,
        mark=diamond*,
        mark options={fill=orange},
        smooth
    ] coordinates {
        (0.05, 1.00)
        (0.1, 1.00)
        (0.2, 1.00)
        (0.4, 0.99)
        (0.6, 0.99)
        (0.8, 0.99)
    };

    \end{axis}
    \end{tikzpicture}
\end{minipage}%
\hfill % Creates horizontal space between the plot and the text
\begin{minipage}{0.4\textwidth} % Right minipage for the text
For each class and sparsity pair, we ran four experiments with different out- and in-shift combinations, and the results for each combination were averaged over 5 runs. %The plot shows the relationship between accuracy and sparsity for different numbers of classes (3, 5, 7, and 9). 
As the sparsity increases, the accuracy remains stable around 1.0 for all classes, with only minor drops at higher sparsity values. 
This indicates that the classification models maintain high performance even with lower data density.
%Variations in accuracy are minimal, especially for  5, 7, and 9 classes, with small fluctuations as sparsity increases. 
\end{minipage}

\caption{Accuracy vs. sparsity for different number of pure classes.} \label{fig:random_plot1}
\end{figure}

\item {\textbf{Mixed Classes:}} For mixed class sets, we used two single-labeled temporal graphs \(G\),
each with 100 vertices, sparsity between 0.05 and 0.8, and time labels in
\((0,100]\). This setup yields both similar and distinct classes. For
instance, Class \(2\text{--}2\) includes two classes from a root graph and
its out-shifted variant, and similarly two from a second root graph.
Classes \(2\text{--}3\) and \(3\text{--}3\) were generated in the same
way. Out-shifts changed on average 4.75\% of interactions
(\(1 \le \epsilon \le 5\)), while in-shifts affected about 1.88\% within
the same range.

%We generate two single labeled temporal graphs $G$ and $G'$  100 vertices and $e$ interactions randomly. We then created $n$ root of graphs $G_1,\dots,G_n$ and $m$ root graphs $G'_1,\dots,G'_m$ using the TP procedure on $G$ and $G'$ respectively. Each class is then created using the TP procedure, but on a smaller percentage of interactions.

%We provide accuracy results for temporal graphs across various sparcity in 
%Tables~\ref{tab:exp3-0.2}, ~\ref{tab:exp3-0.4}, ~\ref{tab:exp3-0.6}, ~\ref{tab:exp3-0.8},
% ~\ref{tab:exp3-0.1} and \ref{tab:exp3-0.05}.

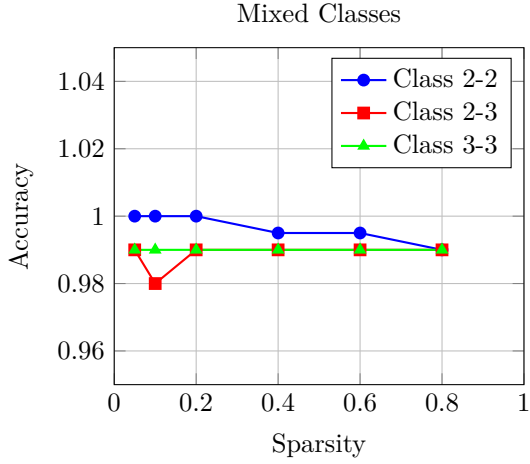
\begin{figure}[hpt]
    \centering
    \begin{minipage}{0.5\textwidth}
        \begin{tikzpicture}
        \begin{axis}[
            width=1.0\textwidth,
            title={Mixed Classes},
            xlabel={Sparsity},
            ylabel={Accuracy},
            xmin=0, xmax=1,
            ymin=0.95, ymax=1.05,
            legend pos=north east,
            grid=both,
            every axis plot/.append style={thick},
            cycle list={
                {blue,mark=*},
                {red,mark=square*},
                {green,mark=triangle*},
                {orange,mark=diamond*}
            }
        ]

        % Data for Class 2-2
        \addplot coordinates {(0.05, 1.00) (0.1, 1.00) (0.2, 1.00) (0.4, 0.995) (0.6, 0.995) (0.8, 0.99)};
        
        % Data for Class 2-3
        \addplot coordinates {(0.05, 0.99) (0.1, 0.98) (0.2, 0.99) (0.4, 0.99) (0.6, 0.99) (0.8, 0.99)};
        
        % Data for Class 3-3
        \addplot coordinates {(0.05, 0.99) (0.1, 0.99) (0.2, 0.99) (0.4, 0.99) (0.6, 0.99) (0.8, 0.99)};
        
        \legend{Class 2-2, Class 2-3, Class 3-3}
        \end{axis}
        \end{tikzpicture}
    \end{minipage}
    \hspace{0.5cm}
    \begin{minipage}{0.35\textwidth}
For each class and sparsity pair, we conducted four experiments with different out- and in-shift combinations, averaging the results over five runs. The observed trends were consistent with the previous experiment (Figure~\ref{fig:random_plot1}), indicating that our classification model performs effectively in the mixed setup as well.

\end{minipage}
    \caption{Accuracy vs. sparsity for different number of mixed classes.}  \label{fig:random_plot2}
\end{figure}

\end{enumerate}

%For real datasets, typically, only a single temporal graph is available, whereas synthetic datasets can provide one or more initial graphs, termed \textbf{root graphs}. Classes are created as described in Section~\ref{sec:Stability}. The TP, EWLSS, and RE procedures generate loosely similar graphs, while the CM procedure produces distinct ones. With a single root graph, two classes are formed: one with loose copies of the root graph and another with multiple CM-generated graphs. For multiple root graphs, classes are formed by loosely copying each root graph.

 \newpage

\subparagraph{Implementation Details:}
All experiments were implemented in Python, with C++ used to compute the
kernel matrices and the \(f_\mathrm{avg}^\mathrm{mlt}\) filtration for
multi-labeled temporal graphs. Experiments were run on a server equipped
with an Intel(R) Core(TM) i7-6950X CPU and dual NVIDIA GeForce RTX 2080
Ti GPUs to ensure consistent runtime and accuracy comparisons.

\subparagraph{Future Work:} The idea of defining a filtration on a temporal graph naturally extends
to higher-order motifs, interval-based temporal graphs,
and probabilistic interactions, opening several avenues for future work
on dynamic systems. With the recent addition of filtered graph kernels to
the toolkit, our methodology further supports the development of
Topological Machine Learning (TML) for temporal data, with potential
applications in epidemic modeling, social networks, and financial
systems.

\section*{Acknowledgement} We thank Priyavrat Deshpande, Writika Sarkar, Mohit Upmanyu, and Shubhankar Varshney for their valuable input during the early discussions of the project.

\section*{Funding} Last three authors received partial support from a grant provided by Fujitsu Limited. Part of this work was carried out during the second author's visit to The University of Newcastle, NSW, Australia, and was partially funded by an ARC Discovery Grant (Grant Number: G2000134).

\newpage
\bibliographystyle{plain}
\bibliography{references}

\end{document}